\documentclass[letterpaper, 10 pt, conference]{ieeeconf}  

\IEEEoverridecommandlockouts                              

\usepackage{tikz}

\usepackage{times}
\usepackage{graphicx}
\DeclareGraphicsExtensions{.pdf,.png,.jpg}
\usepackage{amsmath}
\DeclareMathOperator*{\argmin}{argmin} 
\usepackage{amssymb}  
\usepackage{makecell}
\usepackage{booktabs}
\usepackage{multirow}
\usepackage{here}
\usepackage{kotex}

\title{\LARGE \bf
A Variational Observation Model of 3D Object for Probabilistic Semantic SLAM
}

\author{H. W. Yu and B. H. Lee
\thanks{H. W. Yu and B. H. Lee are with Automation and Systems Research Institute, Department of Electrical and Computer Engineering, 
	Seoul National University, Seoul, Korea(Republic of)
	{\tt\small bgus2000@snu.ac.kr}}%
}

\begin{document}

\maketitle
\thispagestyle{empty}
\pagestyle{empty}

\begin{abstract}
We present a Bayesian object observation model for complete probabilistic semantic SLAM.
Recent studies on object detection and feature extraction have become important for scene understanding and 3D mapping.
However, 3D shape of the object is too complex to formulate the probabilistic observation model;
therefore, performing the Bayesian inference
of the object-oriented features as well as their pose is less considered.
Besides, when the robot equipped with an RGB mono camera only observes the projected single view of an object, a significant amount of the 3D shape information is abandoned.
Due to these limitations, semantic SLAM and viewpoint-independent loop closure using volumetric 3D object shape is challenging.
In order to enable the complete formulation of probabilistic semantic SLAM, we approximate the observation model of a 3D object with a tractable distribution.
We also estimate the variational likelihood from the 2D image of the object to exploit its observed single view.
In order to evaluate the proposed method, we perform pose and feature estimation, and demonstrate that the automatic loop closure works seamlessly without additional loop detector in various environments.
\end{abstract}

\section{INTRODUCTION}
In robotics, simultaneous localization and mapping (SLAM) with high-level features finds various real-world tasks such as autonomous navigation and object search \cite{monocularNavigationAutonomous}\cite{object_search}.
Object-oriented features have evolved as a practical choice for semantic SLAM \cite{slam++}, as they are ideal for view-independent loop closure and complete volumetric maps.
Since mono cameras are popular for SLAM on platforms such as hand-held devices and mobile robots \cite{lsdSLAM}\cite{orbSLAM}, semantic feature detections in single RGB images are widely used.
In particular, thanks to the advent of the neural networks, real-time 2D object detection has been improved and exploited in SLAM implementations.
However, existing object detection methods hardly consider the probabilistic observation model as they mainly focus on the accuracy of the detection and category classification \cite{vggnet}\cite{yolo9000}.
In the case of detection algorithms based on point feature matchings, for example, distinct feature extraction and key point matching are main factors for better performance; therefore, obtaining viewpoint-independent features of the object is relatively less considered \cite{monoObjectSLAM2016PointMatch}.

For viewpoint-independent features, it is ideal to catch the 3D full shape of the object.
The entire shape can be estimated by 3D scanning or SfM \cite{sfmSurvey}, but these methods lack the reality to the mobile platforms that are typically used for various tasks in real-time.
Even if an object-oriented feature is acquired by inferring a 3D shape from a single 2D view \cite{marrnet}\cite{3D_rec_GAN}, landmark pose and feature optimization should be possible for the complete SLAM formulation.
Since the probabilistic distribution of the object with complex 3D shape is intractable, complete SLAM optimization with numerical analysis is still challenging without approximated distributions.
To this end, \cite{yu2018variational} proposes a variational feature encoding method for objects by exploiting their corresponding single view.
However, this method only covers the voxelized 3D single view from depth cameras or range finders, under the assumption of instance segmentation, which is thus insufficient to be directly applied to the real-world scenarios.
The lack of the consideration of object viewpoint orientation also makes it cumbersome to adopt the algorithm when performing the pose optimization in SLAM.

\begin{figure}[t]
	\centering
	\includegraphics[scale=0.075]{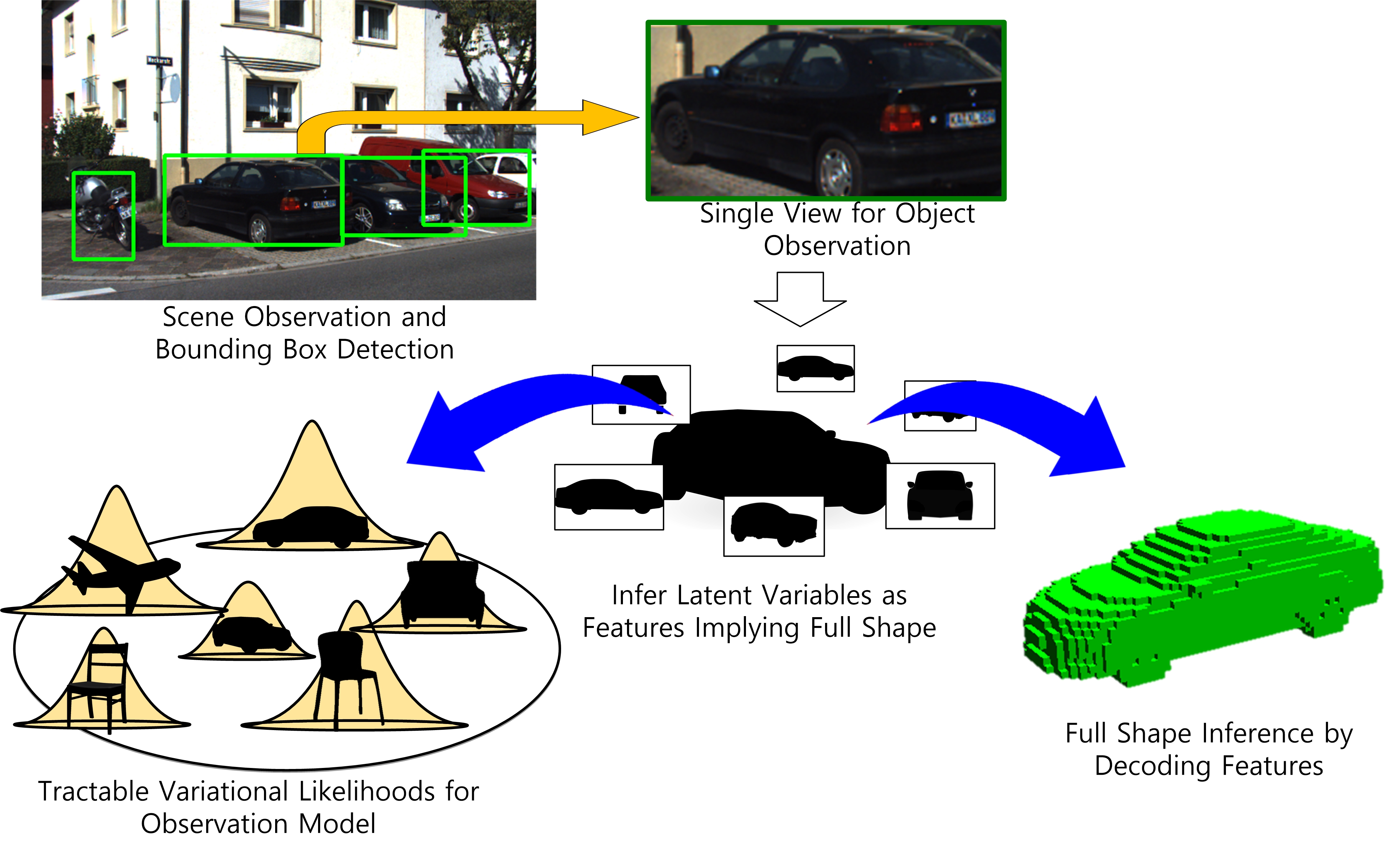}
	\caption{
		In the proposed method, an observed view data is assumed to be an RGB image. In order to approximate the object observation model, we train the variational auto-encoder to achieve the object generative model. The encoded features of the 3D object is estimated from the RGB single view. Using the approximated observation model, complete SLAM formulation and optimization becomes possible.
	}
	\label{likelihood_inference}
\end{figure}
\begin{figure*}[t]
	\centering
	\includegraphics[scale=0.19]{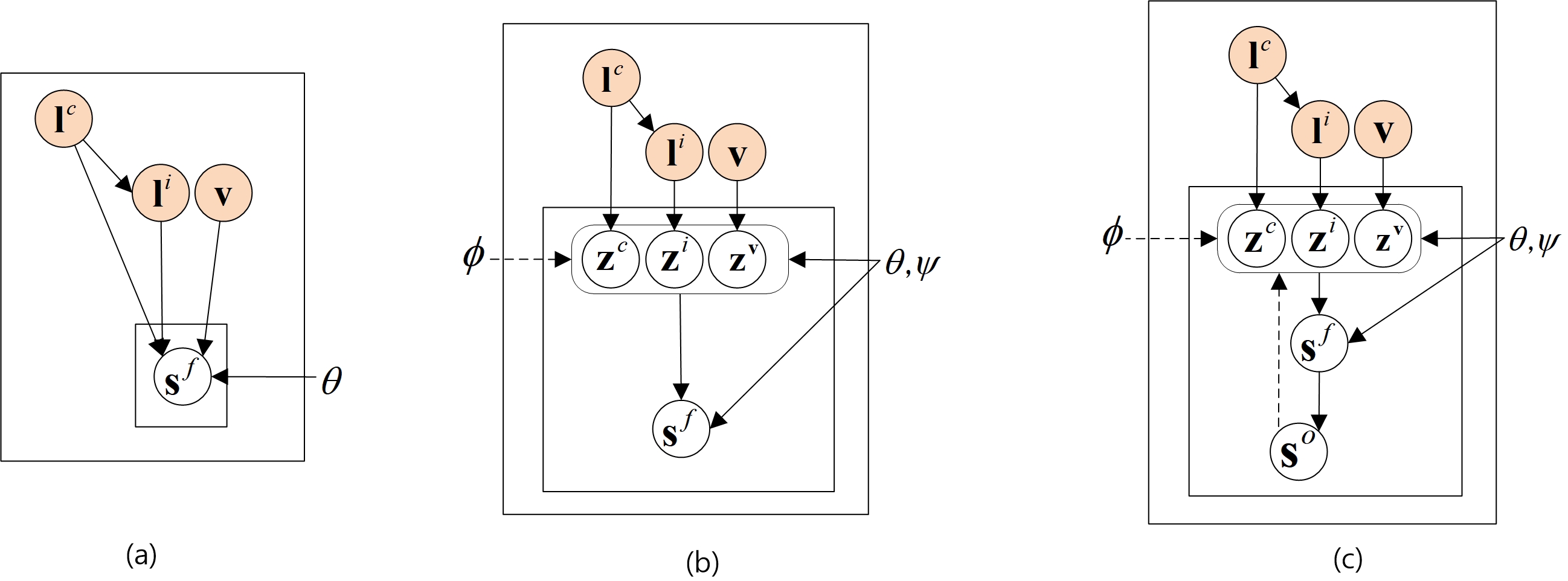}
	\caption{
		Overview of the proposed Bayesian model for the object generative model. (a) The 3D object shape $\boldsymbol{s}^f$ generation is assumed to be involving the category $\boldsymbol{l}^c$, instance $\boldsymbol{l}^i$ and the object orientation $\boldsymbol{v}$ with respect to the observer. $\boldsymbol{v}$ can be modeled with any representation of the orientation such as quaternion or rotation matrix.
		(b) The latent variables according to each element are adopted for the variational likelihood. Priors of $\boldsymbol{l}^c$ and $\boldsymbol{l}^i$ are also learned form the training data by constructing the additional network. Since $\boldsymbol{v}$ is a real value in our case, we simply let $\boldsymbol{v}$ be the mean of the orientation prior.
		(c) The variational likelihood of the proposed observation model is inferred from the observed single view $\boldsymbol{s}^o$. From this strategy, we can exploit the RGB single view without scanning the entire shape of the object directly.
		The generative model is represented by solid lines, and the dashed lines denote the variational likelihood estimation. For the prior, $\psi$ is learned simultaneously with $\theta$ and $\phi$, which are the parameters of the generative model.
	}
	\label{graph_model}
\end{figure*}
Therefore, we propose a method to approximate an object observation model to a tractable distribution, considering the 3D shape and the viewpoint orientation. 
Using the proposed observation model, not only the approximated solutions for the coordinates of the object landmarks but feature values and the orientations can be estimated.
We use the Variational auto-encoder (VAE) \cite{VAE} to infer a variational likelihood of 3D object observation model from RGB single view so that the robot can exploit the observation with RGB mono camera.

Our contributions are two-fold: First, we show that the complete SLAM formulation including feature optimization and automatic loop-closing is available with approximated observation model; second, we propose a real-time encoding method for object observation model with Bayesian generative network, exploiting the RGB single view of the object.

\section{Related Work}
Works on feature-based SLAM typically divide the problem into two parts; front-end for data association and back-end for pose optimization.
Based on the data association using feature matching or tracking algorithm in the front-end, the pose graph optimization is performed at the back-end subsequently.
On semantic SLAM problem, while poses of robot trajectories or of features can be estimated using pose graph optimization methods \cite{g2o}\cite{isam}, obtaining a closed form solution of the optimal feature value is challenging since the high-dimensional features for landmarks follows the intractable distribution.
Due to this complexity, even when object-oriented features are used for classification, features obtained from complex 3D shapes need further learning by using additional algorithms such as SVM \cite{3D_GAN}\cite{ShapeNet}.
Hence, object observation probability is typically considered only on the front-end data association.

Since the object observation model is intractable and hardly considered in SLAM formulation, false-positive data association in the front-end causes tremendous amount of error in the pose graph estimation. 
Besides, the optimal pose is solely estimated without considering the feature values, estimating both robot trajectory, object shape and its pose is challenging.
\cite{probDataSLAM} proposes an expectation maximization (EM) formulation for complete SLAM problem considering data association.
However, as they use the existing object detection method for data association \cite{active_deformable}, it is ambiguous to adopt the object observation probability directly to the SLAM formulation.
Also, they rarely consider the viewpoint-independent feature for 3D shape and object orientation in formulation since they adopt edge and corner filter based deformable part detection algorithm.
In \cite{categorySpecificObjectSLAM}, object pose and shape optimization as well as robot trajectory are estimated in back-end using key point matching and PCA-based object observation factor.
Their observation factor, however, barely considers the tractable object observation model, which is hardly adopt to complete SLAM formulation with computing observation probability.
They are bound to perform pose and shape estimation after data association using additional tracking and loop closure algorithm in the front-end.
\cite{yu2018variational} proposes an encoding method of features which imply the full shape of objects from a single view.
The encoded features follow a tractable Gaussian observation distribution and can be estimated approximately together with robot trajectory and landmark coordinates.
However, the proposed algorithm lacks pragmatic aspects as the single view is limited to the depth image under the assumption that the object is already segmented. 
Also, the viewpoint orientation of the object is hardly considered in SLAM formulation.

We thus introduce a method to object observation model approximation considering the 3D shape and viewpoint orientation.
Using a tractable observation distribution, data association and landmark feature optimization as well as pose optimization are achievable.
Similar to \cite{generate_chairs}, latent variables of class, instance and viewpoint orientation are assumed to be involved in Bayesian graphical model for the object generation. 
In order to exploit the single views obtained as 2D RGB images, we train the network to estimate the variational likelihood of 3D object from 2D single views.
The proposed network is basically VAE, estimated object features can be converted to 3D shapes using decoder as \cite{3D_rec_GAN}\cite{3D_GAN}\cite{generative} to construct a volumetric semantic maps.

\section{Observation Model of Object with 3D Shape and Orientation}

Consider a robot observing the 3D full shape of an object as multiple voxel grids, and using them as a semantic feature $s^f$. Assume that the robot exploits the shape only; not including the other factors such as color, texture or scales of objects.
To obtain the generative model of objects for the approximated observation model, a Bayesian random process can be adopted.
Similar to \cite{generate_chairs}, 
we assume $\boldsymbol{l} = \{\boldsymbol{l}^\omega\}_{\omega=c,i}$ and $\boldsymbol{v}$ cast the Bayesian dice; $\boldsymbol{l}^c$ and $\boldsymbol{l}^i$ stand for the object category and the instance respectively, and $\boldsymbol{v}$ denotes the viewpoint orientation of the object with respect to the observer. We let $v\in\boldsymbol{v}$ be the Euler angle representation.
The graphical model of the proposed process is shown in Fig.~\ref{graph_model}(a).

Now we can derive the joint distribution for the object observation as $p\left(\boldsymbol{s}^f, \boldsymbol{l}, \boldsymbol{v}\right) = p\left(\boldsymbol{l}, \boldsymbol{v}\right)p\left(\boldsymbol{s}^f|\boldsymbol{l}, \boldsymbol{v}\right)$. $p\left(\boldsymbol{l}, \boldsymbol{v}\right)$ is assumed to be uniform; objects in a scene are observed randomly regardless of their category, instance or pose.
To approximate the complex distributions of the object, we can exploit the 
VAE. Variational lower-bound $\mathcal{L}$ of the likelihood $p\left(\boldsymbol{s}^f|\boldsymbol{l}, \boldsymbol{v}\right)$ then can be written as follows:
\begin{align}
	\nonumber
	\mathcal{L}
	\left(
		\theta, \phi, \psi ; \boldsymbol{s}^f, \boldsymbol{l}, \boldsymbol{v}
	\right)
	&=
	-KL
	\left(
		q_\phi
		\left(
			\boldsymbol{z} |\boldsymbol{s}^f
		\right)
		||
		p_\psi
		\left(
			\boldsymbol{z} |\boldsymbol{l,v}
		\right)
	\right)
	\\
	&+
	\mathbb{E}_{\boldsymbol{z}^{\boldsymbol{l,v}}}
	\left[
		\log p_\theta
		\left(
			\boldsymbol{s}^f | \boldsymbol{z}^{\boldsymbol{l,v}}
		\right)
	\right],
	\label{lower_bound}
\end{align}
where $\boldsymbol{z}^{\boldsymbol{l,v}} \sim q_{\phi}\left(\boldsymbol{z}|\boldsymbol{s}^f\right)$.  $\boldsymbol{z}^{\boldsymbol{l,v}}$ can be factorized to $ \boldsymbol{z}^c, \boldsymbol{z}^i, \boldsymbol{z}^{\boldsymbol{v}}$ \cite{VAE}\cite{ladderVAE}, and the KL-divergence in \eqref{lower_bound} is represented as follows:
\begin{align}
	\nonumber
	KL
	\left(
		q_\phi
		\left(
			\boldsymbol{z} |\boldsymbol{s}^f
		\right)
		||
		p_\psi
		\left(
			\boldsymbol{z} |\boldsymbol{l}
		\right)
	\right)
	&=
	\sum_{\omega = c,i}
	KL
	\left(
		q_{\phi^\omega}
		\left(
			\cdot
		\right)
		||
		p_\psi
		\left(
			\boldsymbol{z} | \boldsymbol{l}^\omega
		\right)
	\right)
	\\
	\nonumber
	&+
	KL
	\left(
		q_{{\phi}^{\boldsymbol{v}}}
		\left(
			\cdot
		\right)
		||
		p_\psi
		\left(
			\boldsymbol{z} | \boldsymbol{v}
		\right)
	\right)
	.
\end{align}
We let the prior $p_\psi\left(\boldsymbol{z}|\boldsymbol{l}^{\omega}\right)$ be a Gaussian distribution $\mathcal{N}\left(\boldsymbol{z}; \boldsymbol{\mu}^{\omega}, \boldsymbol{I}\right)$. The corresponding model with latent variable $\boldsymbol{z}$ is displayed in Fig.\ref{graph_model}(b). Here, Gaussians with different $\boldsymbol{\mu}^{\omega}$ should be defined for each label to obtain distinctive features with different shapes. To achieve this, we can construct a prior network for nonlinear function $\boldsymbol{\mu}^{\omega} = f_{\psi}\left(\boldsymbol{l}^{\omega}\right)$ which is trained with VAE simultaneously \cite{yu2018variational}.
The variational likelihood $q_\phi$ is also assumed to be a multivariate Gaussian distribution; $q_{{\phi}^\omega}\left(\cdot\right)= \mathcal{N}\left(\boldsymbol{z}; \boldsymbol{\mu}^{\boldsymbol{s}\omega}, (\boldsymbol{\sigma}^{\boldsymbol{s}\omega})^2\boldsymbol{I}\right)$
and
$q_{{\phi}^{\boldsymbol{v}}}\left(\cdot\right)= \mathcal{N}\left(\boldsymbol{z};\boldsymbol{\mu}^{\boldsymbol{sv}}, (\boldsymbol{\sigma}^{\boldsymbol{sv}})^2\boldsymbol{I}\right)$.
The encoded variables $\boldsymbol{\mu}^{\boldsymbol{s}}=(\boldsymbol{\mu}^{\boldsymbol{s}c}, \boldsymbol{\mu}^{\boldsymbol{s}i})$ then can be viewed as a shape feature, independent to the orientation $\boldsymbol{v}$.
Note that the latent variable $\boldsymbol{z}$ encoded from the variational likelihood $q_\phi$ follows Gaussian priors with $\boldsymbol{\mu}=(\boldsymbol{\mu}^c, \boldsymbol{\mu}^i)$, defined above.

Unlike the category and instance labels $\boldsymbol{l}^c$ and $\boldsymbol{l}^i$, Euler angles $\boldsymbol{v}$ can be directly applied to the Gaussian observation model; therefore we simply let $p_\psi\left(\boldsymbol{z}|\boldsymbol{v}\right) = \mathcal{N}\left(\boldsymbol{z};\boldsymbol{v}, (\sigma^{v})^2\boldsymbol{I}\right)$ with constant $\sigma^{v}$. Unfortunately, Euler angles(both radian and degree) are periodicity, i.e., $0$ and $2\pi$ radians actually represent the same orientation, which makes the network almost impossible to understand the object pose \cite{generate_chairs}. Instead, we let the network to infer $\sin v$ and $\cos v$. 
The prior $p_\psi\left(\boldsymbol{z}|\boldsymbol{v}\right)$ then changes as follows:
\begin{align}
\nonumber
	p_\psi\left(z|\cos v\right)
	\simeq
	\mathcal{N}\left(
		v;
		e^{-(\sigma^v)^2/2}\cos v,
		\sigma^{cos}
	\right),
	\\
	\nonumber
	p_\psi\left(z|\sin v\right)
	\simeq
	\mathcal{N}\left(
		v;
		e^{-(\sigma^v)^2/2}\sin v,
		\sigma^{sin}
	\right)
\end{align}
where $\sigma^{cos} = 1/2+1/2e^{-2(\sigma^v)^2}\cos2v-e^{-(\sigma^v)^2}\cos^2v$,
and
$\sigma^{sin} = 1/2+-1/2e^{-2(\sigma^v)^2}\cos2v-e^{-(\sigma^v)^2}\sin^2v$.
For the simple network structure which commonly expects the diagonal covariance, we assume that $\sin v$ and $\cos v$ are independently sampled. The trigonometric functions are obviously not independent, therefore in actual training we give the additional constraint that $\sin^2v + \cos^2v = 1$. With inferred $\sin v$ and $\cos v$, 
we easily obtain the rotation matrix of $\boldsymbol{v}$ for SLAM that follows the original prior $p_\psi\left(\boldsymbol{z}|\boldsymbol{v}\right)$.

Since the actual observation of objects mainly comes as the form of an RGB single view, we redefine the variational likelihood as $q_\phi\left(\boldsymbol{z}|\boldsymbol{s}^o\right)$, where $\boldsymbol{s}^o$ is a single view image of the object \cite{yu2018variational}\cite{3D_GAN}. The Bayesian graph model involving $\boldsymbol{s}^o$ is shown in Fig.\ref{graph_model}(c).

\section{Variational Latent Features and Semantic SLAM}
\subsection{Variational Features of Shape and Orientation for EM Formulation}
Suppose a mobile robot navigates the unknown area up to time step $T$. Let robot poses and a set of object observations be $\mathcal{X}=\{\boldsymbol{x}_t\}^T_{t=1}$ and $\mathcal{S}=\{\mathcal{S}_t\}^T_{t=1}$, respectively.
$k$'th object detection 
$\boldsymbol{s}_k 
= 
(
\boldsymbol{s}_k^f, \boldsymbol{s}_k^{\boldsymbol{t}}
) \in \mathcal{S}_t$
on time step $t$ is composed of a full shape $\boldsymbol{s}^f$ and  a 3D coordinate $\boldsymbol{s}^{\boldsymbol{t}}$. 
Similar to \cite{probDataSLAM}, assume that we previously have static landmarks $\mathcal{L}=\{l_m=\left(l^p_m, l^c_m\right) \}^M_{m=1}$, where $l^p=\{\boldsymbol{t},\boldsymbol{v}\}$ denotes the landmark pose composed of translation $\boldsymbol{t}$ and orientation $\boldsymbol{v}$. The object label $l^c$ consists of category and instance; $l^c=\boldsymbol{l}=\{\boldsymbol{l}^c, \boldsymbol{l}^i\}$.
We let $\mathbb{D}_t$ be the set of all possible data associations 
$\mathcal{D}_t = \{\left(\alpha_k, \beta_k \right) \}^K_{k=1}$ which denotes that the object detection $\boldsymbol{s}_k$ of landmark $l_{\beta_k}$ was obtained at the robot pose $\boldsymbol{x}_{\alpha_k}$.
Also,
$\mathbb{D}_t\left(i,j\right)\subseteq\mathbb{D}_t$ is the set of all possible data association $\mathcal{D}_t^\prime = \{\left(\alpha_k^\prime, \beta_k^\prime \right) \}$ representing that $i$th object detection is assigned to $j$th landmark.

Now assume that observation events are iid, and we have
\begin{align}
\nonumber
	p\left(\mathcal{S}|\mathcal{X},\mathcal{L},\mathcal{D}\right)
	= 
	\prod_{k}
	p(
	\boldsymbol{s}_k^t|\boldsymbol{x}_{\alpha_k}, \boldsymbol{t}_{\beta_k}
	)
	p(
	\boldsymbol{s}_k^{f}|\boldsymbol{x}_{\alpha_k}, \boldsymbol{v}_{\beta_k}
	)
	p(
	\boldsymbol{s}_k^{f}|\boldsymbol{l}_{\beta_k}
	).
\end{align}
Similar to \cite{yu2018variational}, using variational lower bound in \eqref{lower_bound} to approximate the observation model $p\left(\boldsymbol{s}^f|\boldsymbol{l},\boldsymbol{v}\right)$, the EM formulation for the complete semantic SLAM can be derived as follows:
\begin{align}
\label{emWeight}
	w^t_{ij}
	&\simeq
	\frac
	{
		\sum_{\mathcal{{D}}_t^\prime \in \mathbb{D}_t\left(i,j\right)}
		\prod_{k}
		p\left(
		\boldsymbol{s}^p_k|\boldsymbol{x}_{\alpha_k^\prime},l^p_{\beta_k^\prime}
		\right)		
		p_\psi\left(
		\boldsymbol{\mu}^{\boldsymbol{s}}_k
		|
		\boldsymbol{l}_{{\beta}^\prime_k}
		\right)
	}
	{
		\sum_{\mathcal{{D}}_t \in \mathbb{D}_t}
		\prod_{k}
		p\left(
		\boldsymbol{s}^p_k|\boldsymbol{x}_{\alpha_k},l^p_{{\beta}_k}
		\right)
		p_\psi\left(
		\boldsymbol{\mu}^{\boldsymbol{s}}_k
		|
		\boldsymbol{l}_{{\beta}_k}
		\right)
	}
	\\
	\label{poseOptimize}
	\mathcal{X}, l^p
	&=
	\argmin_{\mathcal{X}, l^p}
		\sum_{t,k,j}
			-w^t_{kj}
			\log 
				p\left(\boldsymbol{s}^{\boldsymbol{t}}_k|\boldsymbol{x}_t, \boldsymbol{t}_j\right)
				p\left(\boldsymbol{\mu}^{\boldsymbol{sv}}_k|\boldsymbol{x}_t, \boldsymbol{v}_j\right),
	\\
	\label{shapeOptimize}
	\boldsymbol{l}
	&=
	\argmin_{l^c}
	\sum_{t,k,j}
	-w^t_{kj}
	\log
	p_\psi\left(\boldsymbol{\mu}^{\boldsymbol{s}}_k | \boldsymbol{l}_j \right),
\end{align}
where 
$p_\psi\left(\boldsymbol{z}|\boldsymbol{l}\right)
= 
p_\psi(\boldsymbol{z}|\boldsymbol{l}^c)
p_\psi(\boldsymbol{z}|\boldsymbol{l}^i)
$.
For more details of the approximated EM formulation with variational lower bound, please refer to \cite{yu2018variational}.
Therefore, the approximated solution considering the 3D shape and its orientation of the semantic SLAM can be obtained, even if only the RGB single view $\boldsymbol{s}^o$ of objects is observed.
\begin{figure*}[t]
	\centering
	\includegraphics[scale=0.15]{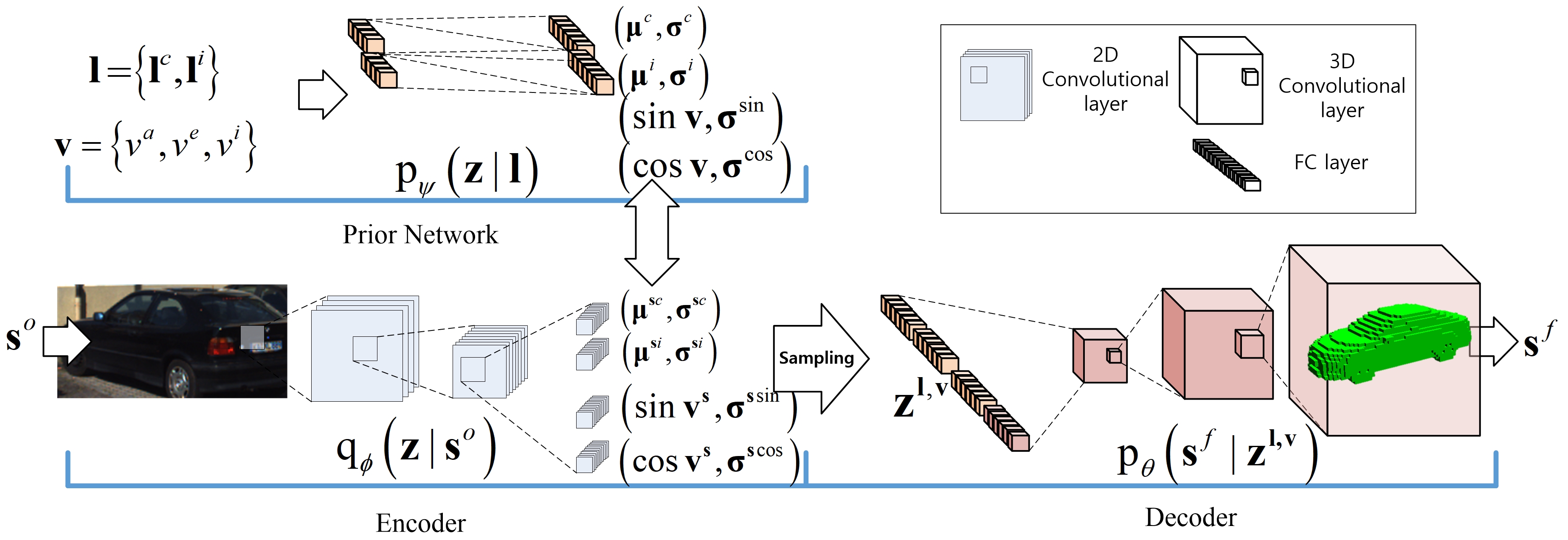}
	\caption{
		Proposed network architecture.
		The encoder part for the variational likelihood estimation is constructed using darknet-19 which is used in YOLOv2. We add an additional 2D convolution layer at the end of the darknet-19 so that the encoder becomes a fully convolution network.
		The decoder is composed of a dense layer and several 3D convolution-transpose layers. The prior networks which is trained with VAE simultaneously consist of fully connected layers. The end-to-end training of the proposed networks is achievable.
	}
	\label{proposed_networks}
\end{figure*}

\subsection{Pose and Feature Optimization of 3D Object}

For the case of landmarks, object-oriented SLAM basically deals with finite static targets \cite{slam++}\cite{probDataSLAM}\cite{categorySpecificObjectSLAM}; therefore, to get optimal labels $l^c=\boldsymbol{l}=(\boldsymbol{l}^c, \boldsymbol{l}^i)$ of \eqref{shapeOptimize}, it is necessary to perform maximum likelihood estimation (MLE) by substituting the all static landmark labels that are previously known.
However, since we adopt the VAE which is a probabilistic generative model, we can expand the static and discrete landmark features(denoted as labels) to the continuous ones(as latent variables). Although the network learns from the finite datasets and observes the finite 3D shapes, it can take advantage of nonlinear regression which fits the encoded latent variables into the continuous latent space. \cite{GAN},  \cite{DCGAN} and \cite{3D_GAN} also show that `walking in the latent space' of 2D images or 3D objects would be available. In the same vein, we can assume the continuous feature observation model. 
In other words, the prior of $\boldsymbol{z}$ in \eqref{shapeOptimize} is defined not as $p_\psi\left(\boldsymbol{z}|\boldsymbol{l}\right)$, but as $p_\psi\left(\boldsymbol{z}|\boldsymbol{\mu}\right)$.
We therefore find not the optimal labels $\boldsymbol{l}$ of the object, but optimal shape features $\boldsymbol{\mu}=(\boldsymbol{\mu}^c, \boldsymbol{\mu}^i)$. To achieve this task, we simply derivate \eqref{shapeOptimize} which is a weighted sum of the logarithm of Gaussian. The optimal $\boldsymbol{\mu}$ then can be represented as the following:
\begin{align}
	\boldsymbol{\mu}_j = \sum_{t,k,j} -w^t_{kj} \boldsymbol{\mu}^{\boldsymbol{s}C}_k
	\label{optimalShape}
\end{align}
Consequently, without considering the static landmarks used in training, the optimal shape feature of $j$th observation is easily computed by replacing \eqref{shapeOptimize} with \eqref{optimalShape}.
For the pose estimation of \eqref{poseOptimize}, the optimal pose can be calculated with nonlinear pose graph optimizer such as \cite{g2o}\cite{isam}. 

After iterating \eqref{emWeight} for \eqref{shapeOptimize} for EM algorithm, we can obtain the approximated solutions of not only the poses of the robot trajectory and the objects, but also the optimized features of 3D shapes. The volumetric full shapes of objects also can be obtained by decoding 
$\boldsymbol{\mu}_j$.

\section{Training Details}
\subsection{Training Datasets and Data Augmentation}
The proposed network is assumed to have a single image that contains exactly one object as an input.
In order to train this network, we use 2D object images from PASCAL3D \cite{pascal3D} and ObjectNet3D \cite{objectnet3d} dataset cropped by the ground truth bounding box. The annotated 3D shapes are also used as the ground truth 3D shape, after being converted to the voxelized grid form.
We arbitrary choose 40 object categories in practice from the training datasets, which contain almost 100 categories in total. All instances in each class are fully used.
While training, we use data augmentation for 2D images to avoid overfitting similar to \cite{yolo9000}. Since our network should infer the orientation of the 3D object from the 2D image, random rotation and flipping are discarded which can change the object orientation.

\begin{figure*}[t]
	\centering
	\includegraphics[scale=0.25]{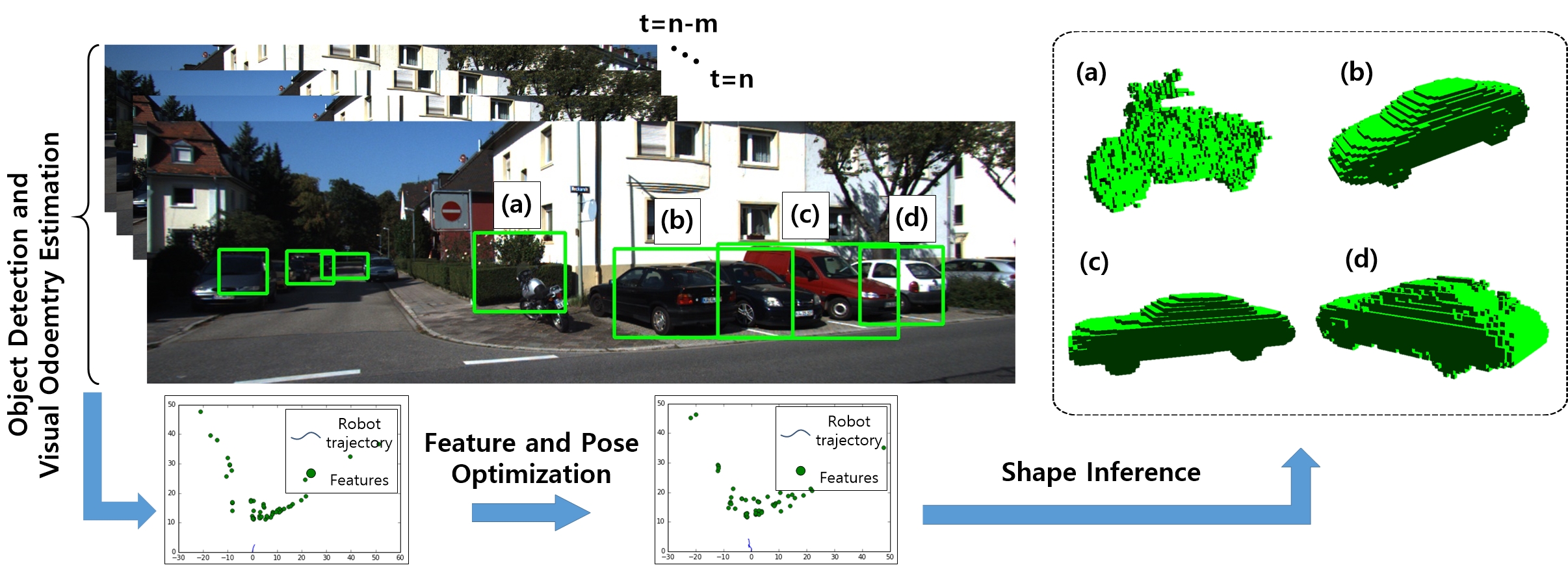}
	\caption{
		Overview of the proposed system.
		Basically we estimate the visual odometry for every sequences.
		For each of the keyframes, object detection and feature encoding are performed simultaneously.
		Using the encoded features and the approximated observation model, SLAM optimization considering the object shape and the orientation is conducted.
		To solve the complete SLAM formulation, only the encoding process using the trained encoder is necessary.
		However, the full shape reconstruction of each observed object can be obtained at anytime.
	}
	\label{trjAndShape}
\end{figure*}
\subsection{Loss Function}
The negative variational lower bound from \eqref{lower_bound} is defined as the VAE loss \cite{VAE}\cite{3D_GAN}. The network is basically an auto-encoder, thus the expectation term in \eqref{lower_bound} can be regarded as the shape reconstruction loss $L^{rc}$. In our encoding scheme, network learns various tasks: encoding the shape features of category as well as instance level, and the orientations represented in trigonometric form. Since the network tries to deal with these tasks simultaneously, it hardly reaches the optimal point especially for the shape reconstruction. To prevent the network from plunging into the local minimum, we penalize the false negatives of the predicted voxel grids by defining $L^{rc}$ as follows \cite{generative}:
\begin{align}
\nonumber
	&L^{rc} = 
	\\
	\nonumber
	&\sum_j 
	-\gamma \boldsymbol{s}_j^{ft} \log\left(\boldsymbol{s}_j^{fp}\right)
	-\left(1-\gamma\right)\left(1-\boldsymbol{s}_j^{ft}\right)\log \left(1-\boldsymbol{s}_j^{fp}\right),
\end{align}
where $\boldsymbol{s}_j^{fp}$ and $\boldsymbol{s}_j^{ft}$ are the $j$th component of the binary variables for the predicted and the target 3D shape $\boldsymbol{s}^f$, respectively. By setting $\gamma$ up to $0.5$, our network avoid the local minimum that simply predicts the empty voxel grids.

For the pose estimation of the object shape, we choose the Euler angle composed of azimuth, elevation and in-plane rotation angles in radian. Since the deviation $\sigma^v$ of each angle for the approximated observation model can be manually selected, we set $\sigma^v = 0.05\text{}rad \simeq 2.87^{\circ}$.

\subsection{Networks}
Recent researches of the monocular SLAM typically aim the real-time performance.
To meet this trend, we adopt darknet-19 for our encoder, which is capable of real-time operation \cite{yolo9000}.
The encoder is then constructed by adding a convolution layer followed by a global max-pooling layer on top of the darknet-19 network.
When the encoder is solely used for feature encoding except the decoder for 3D shape reconstruction, it can operate at about $30$ fps.
In practice, basically 10 to 15 objects are observed in a single scene; when the images of objects are cropped from a scene go parallel to the encoder, the encoding process still can operate at about $30$ fps.
Since our network learns about the objects and their pose from the 2D RGB images, we pre-train the network for two tasks in order; object category classification of the Imagenet dataset \cite{imagenet}, and orientation classification of the Render for CNN dataset \cite{renderforcnn}.

The decoder of our network follows the structure introduced in \cite{3D_GAN}.
To make the decoder familiar with the 3D object shape, we construct the VAE using an additional 3D encoder together, and pre-train the network using ModelNet40 dataset \cite{ShapeNet}.
In this pre-training of the decoder, a prior network consists of $3$ dense layers are pre-trained together.
After pre-trainings
, we combine these networks and train the proposed observation model. The proposed network structure is displayed in Fig.~\ref{proposed_networks}.

\begin{figure*}[t]
	\centering
	\includegraphics[scale=0.4]{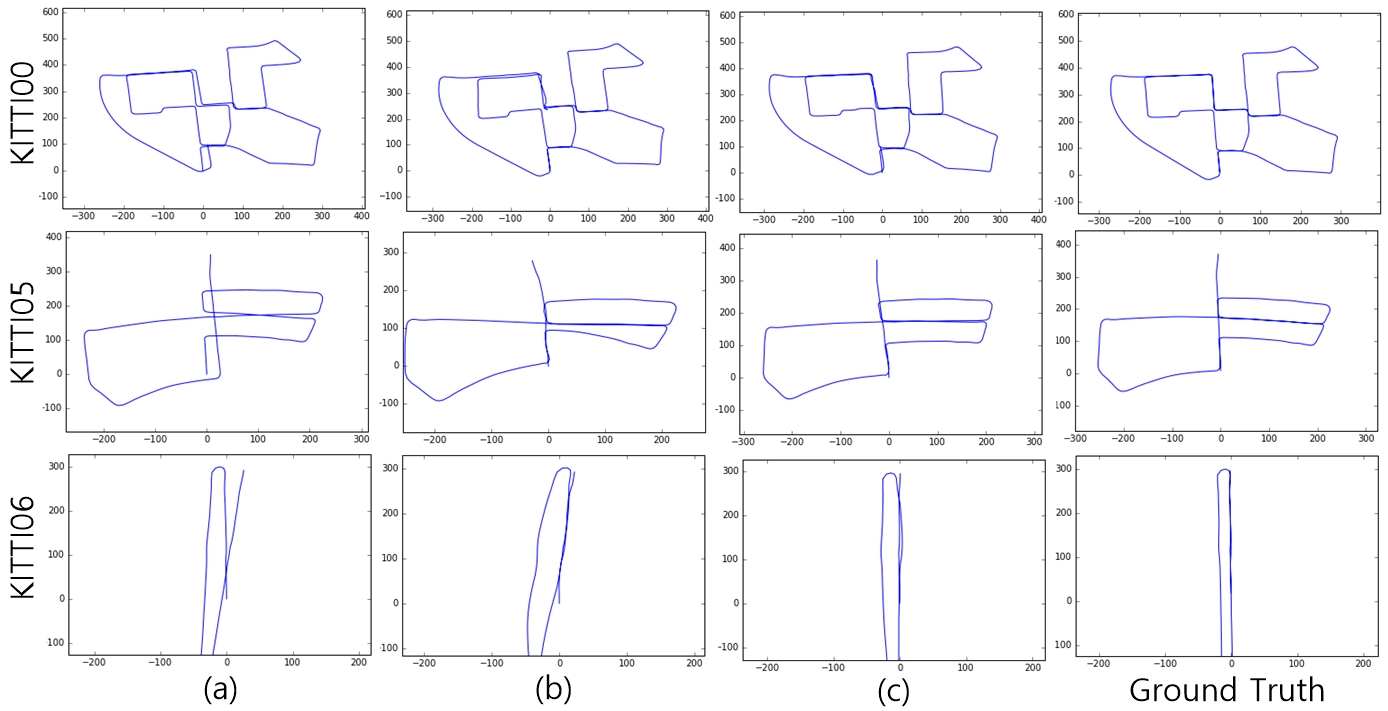}
	\caption{
		Comparison of the estimated robot trajectory from SLAM. From first to last rows: results of the KITTI 00, 05 and 06 sequences respectively. (a-c): visual odometry, estimation results with shape features only, and results with shape and orientation together, respectively.
		Since most of the objects in KITTI dataset are car category and have similar shapes to each other, SLAM with only shape feature hardly performs the complete loop closing.
		Results with features obtained from imagenet pre-trained darknet=19 and vggnet are not displayed as they have no significant difference to the visual odometry results.
	}
	\label{trjCompare}
\end{figure*}
\begin{table*}[h]
	\caption{Comparison of the Robot Trajectory Estimation}
	\label{ATEandRPE}
	\begin{center}
		\begin{tabular}{c c c c c c c c c c c c}
			\hline
			Encoding & \multicolumn{2}{c}{KITTI00} & &\multicolumn{2}{c}{KITTI05} & &\multicolumn{2}{c}{KITTI06} & &\multicolumn{2}{c}{TUMf3} \\
			\cline{2-3}\cline{5-6}\cline{8-9}\cline{11-12}
			Methods & ATE(m) & RPE(m) & & ATE(m) & RPE(m) & & ATE(m) & RPE(m) & & ATE(m) & RPE(m)\\
			\hline
			visual odometry &  234.94 & 249.21 & & 33.52 & 44.28 & & 28.11 & 31.56 & & 3.00 & 2.90 \\
			ours(w/o object orientation) & 44.34 & 46.82 & & 43.07 & 39.69 & & 33.40 & 30.69 & & 2.87 & 2.97 \\
			ours(with object orientation) & 31.55 & 29.33 & & 25.11 & 29.45 & & 20.83 & 18.48 & & 2.87 & 2.86 \\
			\hline
		\end{tabular}
	\end{center}
\end{table*}
\section{Experiments}
\subsection{Object Detection and Feature Encoding}
In order to achieve the real-time object detection and obtain the bounding boxes of objects in 2D scene with multiple objects, we adopt YOLOv2 \cite{yolo9000}. 
Object images obtained by cropping bounding boxes in the scene are passed through our network in parallel.
Since the encoder is solely used for the feature encoding, it takes almost twice as much time as \cite{yolo9000} to perform the object detection and feature encoding altogether.
For the efficiency of SLAM, we simply select every 15th frame as a keyframe where detection and feature encoding is performed. 

\subsection{Pose and Feature Graph Optimization}
For the initial guess of robot trajectory $\mathcal{X}$ in \eqref{poseOptimize}, we use visual odometry based on 8 point algorithm \cite{visualOdometry}. Additional optimizations for odometry correction such as local bundle adjustment are not used.
To avoid the scale problem, similar to \cite{probDataSLAM} we assume that the real-world scale can be known from the device such as IMU. Note that we exploit the IMU data not for the robot pose estimation, but only for the real-world scale prediction. The centers of the detected objects are defined with 3D coordinates, given by the median values of each coordinates of the visual features within the object’s bounding box.

In the pose and feature graph, the robot node contains $SE3$ pose $\boldsymbol{x}$. The object node also contains $SE3$ pose $\left(\boldsymbol{t}, \boldsymbol{v} \right)$, and its encoded feature $\boldsymbol{\mu^s}$.
For the edges between features, all observed object pairs are fully connected since all the possible data association should be considered in \eqref{emWeight}. In practice for computational reasons, however, we disconnect the edges between two observed objects when the weight $w$ between them is under the threshold $\delta$.

\subsection{Optimization Results for SLAM}
We use KITTI \cite{kitti} and TUM \cite{tumDataset} datasets for SLAM experiments in various environments.
The object images obtained by the object detection are encoded to features with the proposed method.
For the evaluation of the inferred shape features and the orientations of objects, we compare the optimization results using the aspects above with the results using shape features only.
In order to compare SLAM results with other feature extraction methods, we use three different encoders: encoder for 2D-3D autoencoder, darknet-19 \cite{yolo9000} pretrained for imagenet \cite{imagenet} classification, and vggnet16 \cite{vggnet}.
We train the 2D-3D autoencoder to infer the 3D shape from the 2D image of the object similar to the proposed network, but without any other probabilistic constraints.

The optimization results for the robot trajectory are shown in Fig.~\ref{trjCompare}. Although any additional loop closing techniques are not adopted, the proposed method naturally consider loop closing points in a large space in the SLAM optimization process.
The optimizations with the shape features without the orientations of the objects, failed in places with a lot of vehicles having similar shapes.
Since the obtained features from the 2D-3D autoencoder, pretrained darknet-19 and vggnet follow non-linear distributions, optimization results with these features have no difference with the odometry.
The absolute position errors with respect to ground truth are shown in TABLE \ref{ATEandRPE}. For TUMf3 datasets, which is a sequence of images mainly focusing on a desk, complete optimization is unavailable since our method exploits various objects in the images.

3D reconstructions for the observed objects can be performed after the pose and feature optimization.
After performing SLAM on the KITTI00 sequences, we obtain the 3D shape of the corresponding features with the trained decoder. The reconstruction results for an arbitrary scene in this sequence is shown in Fig.~\ref{trjAndShape}.

\section{CONCLUSION}
We have shown that a variational object observation model enables the complete probabilistic semantic SLAM.
The high-dimensional features such as the 3D shape of objects follow an intractable distribution; therefore in the existing SLAM formulation it is scarcely considered to the optimize the feature as well as the poses of the object.
To overcome this problem, we approximate the observation model of the 3D object to a tractable distribution using a generative model.
Since the robot with a mono camera basically observes an RGB single view of an object, the proposed algorithm infers a variational likelihood of the full shape from the single view.
Pose and feature optimization can be simultaneously performed using the encoded features, and the 3D full shape can also be obtained through the decoding process. 
Our experiments suggest that the graph optimization and automatic loop closing is performed smoothly. 
To evaluate the approximated observation model and the encoded features, we conduct SLAM and shape reconstruction for the environments with various objects.


\section*{Acknowledgement}
We would like to express our gratitude to Jihoon Moon for his comments and proofreading on this manuscript.





\bibliographystyle{IEEEtran}
\bibliography{root}

\end{document}